\def\year{2021}\relax
\documentclass[letterpaper]{article} % DO NOT CHANGE THIS
\usepackage{aaai21}  % DO NOT CHANGE THIS
\usepackage{times}  % DO NOT CHANGE THIS
\usepackage{helvet} % DO NOT CHANGE THIS
\usepackage{courier}  % DO NOT CHANGE THIS
\usepackage[hyphens]{url}  % DO NOT CHANGE THIS
\usepackage{graphicx} % DO NOT CHANGE THIS
\usepackage{natbib}  % DO NOT CHANGE THIS AND DO NOT ADD ANY OPTIONS TO IT
\usepackage{caption} % DO NOT CHANGE THIS AND DO NOT ADD ANY OPTIONS TO IT
\usepackage{microtype}
\usepackage{enumitem}
\usepackage{amssymb}
\usepackage{graphicx}
\usepackage{subcaption}
\usepackage{amsmath}
\usepackage{caption}
\usepackage{lipsum}

\newcommand\oursys{C2PO}
\usepackage[switch]{lineno}
\title{Automated Storytelling via Causal, Commonsense Plot Ordering}
\author{Prithviraj Ammanabrolu, Wesley Cheung, William Broniec, and Mark O. Riedl\\}
\affiliations{
  School of Interactive Computing \\
  Georgia Institute of Technology\\
  \texttt{\{raj.ammanabrolu,wcheung8,d.tu,wbroniec3,riedl\}@gatech.edu}
}
\begin{document}

\maketitle
%\linenumbers
\begin{abstract}
Automated story plot generation is the task of generating a coherent sequence of plot events.
Causal relations between plot events are believed to increase the perception of story and plot coherence. 
In this work, we introduce the concept of {\em soft causal relations} as causal relations inferred from commonsense reasoning.
We demonstrate C2PO, an approach to narrative generation that operationalizes this concept through \textbf{C}ausal, \textbf{C}ommonsense \textbf{P}lot \textbf{O}rdering.
Using human-participant protocols, we evaluate our system against baseline systems with different commonsense reasoning approaches and inductive biases to determine the role of soft causal relations in perceived story quality.
Through these studies we also probe the interplay of how changes in commonsense norms across storytelling genres affect perceptions of story quality.\footnote{Code found at \url{https://github.com/rajammanabrolu/C2PO}.}
\end{abstract}

\section{Introduction}
%why stories are interesting in terms of language

%how we are gonna look at one specific aspect

%stories can be structured into plot graphs, DAGs, this is an indication of causal structure

%how can you generate these things, taking into account commonsense reasoning

Automated story generation is a standing grand challenge of AI.
One of the central challenges of automated story generation is causal progression such that the events of the story follow from events that have come before.
Many prior approaches to plot generation relied on symbolic planning~\cite{Lebowitz1987,Gervas2005,Porteous2009,Riedl2010a,ware11}---reasoning directly about causal enablement in the form of predicate precondition and post-condition matching.
While these systems can guarantee causal entailment between story events, these approaches also require extensive domain knowledge engineering and limited vocabularies of events and characters.

Machine learning approaches to automated story generation can learn storytelling and domain knowledge from a corpus of existing stories or plot summaries. 
This theoretically allows them to overcome the knowledge engineering bottlenecks. 
However, neural language model based approaches to automated story generation learn probabilistic relationships between words, sentences, and events and thus have difficulty modeling causal entailment between actions and events.
Additionally, stories need to remain consistent with respect to genre and commonsense norms.

In this paper, we
consider the challenge of 
%present a novel technique for 
automatically generating narratives that have recognizable causal entailment between events.
%\Mark{Should I distinguish between stories and plots? EMNLP might be happier if we call it plot generation.}
Specifically, we approach the problem of story generation as a {\em plot-infilling}~\cite{ippolito-etal-2019-unsupervised,donahue2020enabling} where an outline of plot points is extracted from a source then elaborated upon. 
We introduce the concept of {\em soft causal relations}, where causal entailment between story events does not need to be strictly logically consistent, but draws upon people's everyday commonsense understanding of whether one event tends to be preceded or succeeded by another.

%Automated storytelling is a standing grand challenge of AI, bringing together problems faced in multiple natural language tasks.
% It requires creative, long-form text generation.
% Stories must be interesting and coherent, modeling long range dependencies in terms of characters and setting and being grammatical short term.
% Additionally, they need to remain consistent with respect to genre and commonsense norms.
% It is this latter aspect that we specifically focus on in this work. \TODO{citeeeeeeeeeee}
% We are motivated by the question of how causality and commonsense reasoning affect perceived story quality in terms of all other human dependent metrics.
% % In an attempt to isolate this aspect of storytelling, we design a system 

%The first question when measuring these aspects of storytelling is how to represent stories so that we can more effectively capture causality.

We demonstrate an approach to story generation using soft causal relations in the \oursys{} (Commonsense, Causal Plot Ordering) system, which generates narratives via plot infilling using soft causal relations. 
Inspired by work on plot graph learning~\cite{Li}, \oursys{} attempts to create a branching space of possible story continuations that bridge between plot points that are automatically extracted from existing natural language plot summaries.
To create this branching story space, we iteratively extract commonsense causal inferences from the COMET~\cite{Bosselut2019COMETCT} model of commonsense reasoning. 
Finally, once the space---a plot graph---has been constructed, we search the space for complete sequences.

Using human participation studies, we evaluate \oursys{} against baseline text infilling systems with different uses of commonsense reasoning and inductive biases to determine the role of soft causal relations on perceptions of story quality.
We choose two story corpora in different genres: real-world mystery stories such as Sherlock Holmes---known for generally being consistent with everyday commonsense norms, and children's fairy tales such as Hansel and Gretel---stories which usually shatter commonsense expectations.
Through these studies we further explore the broader issue of how the change in commonsense norms across storytelling genres affects perceptions of story quality.

\section{Background and Related Work}
%https://arxiv.org/pdf/2001.05139.pdf guan

%We focus on two specific areas of prior work: general approaches to storytelling, and methods that explicitly attempt to incorporate commonsense knowledge into story generation.

Narrative generation systems that use symbolic planning~\cite{Lebowitz1987,Gervas2005,Porteous2009,Riedl2010a,ware11}
explicitly ensure causal relations between actions via predicate calculus operations over explicitly modeled action preconditions and post-conditions.
These symbolic proposition represent {\em hard} causal relations.

Neural language-model based approaches to story generation have typically overlooked causality or assumed it would emerge in the hidden state of neural networks.
\citet{Roemmele2018b} use LSTMs with skip-thought vector embeddings~\citep{kiros2015skip} to generate stories. 
Similarly %\citet{Khalifa2017} train recurrent neural nets on single-author corpora.
\citet{clark-etal-2018-neural}
\citet{Martin2017a,Martin2018} introduce semantic event abstractions known as events and decompose storytelling into the problems of generating event sequence and elaborating the events into natural language.
\citet{Tambwekar2019} extends this work by fine-tuning language models to achieve a given goal, though goals are not necessarily achieved in a causality-preserving way as in symbolic planning.
\citet{Fan2018} and \citet{ammanabrolu2019story} pursue hierarchical approaches to story generation, wherein a prompt is first generated and then transformed into a text passage.
\citet{Yao} break down the problem of story generation into that of planning
%\Mark{is there another word we can use, or does it actually do planning?} - the paper is Plan-And-Write and actually plans looks like
out a story and then generating from it.
%\citet{ammanabrolu2019toward} look at narrative generation as a form of quest generation in interactive fiction and use a knowledge graph to ground their generative models.

\citet{ippolito-etal-2019-unsupervised} look at filling in missing parts from a story by conditioning a text generator on rare words, also attempting to achieve balance between novelty and coherence.
\citet{donahue2020enabling} also attempt to model storytelling along these lines, training a language model to fill in the blanks given left and right contexts.
None of these methods explicitly incorporate commonsense knowledge into story generation.

An alternative machine learning based approach to story generation introduced by~\citet{Li} is to first learn a {\em plot graph} that can then be used as a constrained search space for a sequence of story events.
Plot graphs are directed acyclic dependency graphs where each node represents a plot point or event and the arcs between nodes represent {\em temporal} constraints.
%Formally, a plot graph is represented as $G=\langle E,P,R\rangle$ where $E$ is the set of all plot events and $R=\{(x,y):x\to y; x,y\in E\}$\Mark{What is P? Was expecting $<E,R,E>$. Could probably just cut this formalism} is the ordered set of pairwise event dependencies~\citep{li}.
%
Inspired by this approach, we also attempt to learn a branching story graph structure that can be searched;
however, instead of learning the plot graphs from a crowdsourced text corpus, we construct this graph by extracting commonsense inferences about causally related events.

Approaches to automated story generation that incorporate commonsense resources include the following.
\citet{rashkin-etal-2018-modeling} present an annotation framework specifically designed to examine the mental states of characters in commonsense based stories.
\citet{Guan2019} incorporate external commonsense knowledge sources to explicitly improve story ending generation and \citet{mao-etal-2019-improving,guan2020knowledge} look at fine tuning pre-trained transformer based language models~\cite{vaswani2017attention} on commonsense sources like ConceptNet~\cite{Speer2012} and the BookCorpus~\cite{kiros2015skip}.
These works, however, focus on improving what they call logicality and grammaticality, translating largely to local coherence, as opposed to analyzing perceptions of causality or overall story quality.

\begin{figure*}[t]
    \centering
    \includegraphics[width=0.9\linewidth]{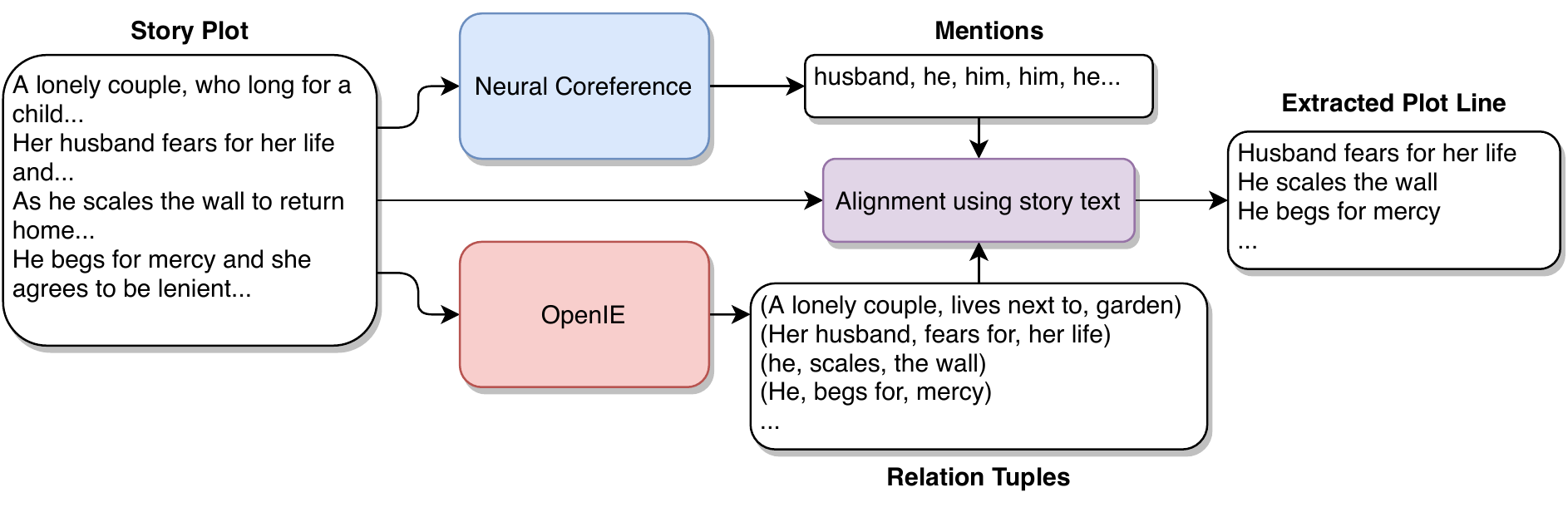}
    \caption{An illustration of high level plot point extraction.}
    \label{fig:extraction}
\end{figure*}

\section{Soft Causal Relations}
\label{sec:causal}

A {\em hard causal relation} implies that some world state transitions that are illegal---e.g., a character John cannot shoot Xavier if John is not in possession of a gun and the two characters are physically co-located.
In contrast,
a {\em soft causal relation} is mediated by the assumed reader's beliefs.
Soft causality is therefore causality--normally a logical construct in narrative--mediated by the beliefs of the reader. 
It provides a causal ordering of events from the perspective of the reader instead of from the perspective of the author (whether human or agent).
That is, a soft causal relation is a reasonable expectation of two non-mutually exclusive criteria: 
(a)~certain activities are needed to achieve a character's goal, and 
(b)~certain activities are in pursuit of future goals.
The first clause draws on the psychological theory of the role of causality in story understanding by \citet{trabasso85}:
readers attempt to understand ``why'' events occur by tracking causal relations as {\em enablement}---some event $y$ cannot occur unless some preceding event $x$ occurred.
The second clause draws upon a theory of the role of character goal hierarchies in story understanding by \citet{graesser91}: readers attempt to understand ``why'' things happen by tracking and predicting character goal hierarchies.
In both cases, whether an inference is made by reader is strongly dependent on what the reader's beliefs about the world are.
In short, the key difference between hard and soft causality is the idea of expectations of causality via commonsense reasoning.

%Further, soft causality differs from local coherence in a couple of ways.
%A closely related concept to soft causality is that of local coherence~\citep{purdy2018predicting}.
%Here, soft causality differs from local coherence in a couple of ways.
%Most notably, soft causality can be a long-range relationship between events, whereas local coherence usually means a thematic relationship between adjacent (or nearby) sentences. 
%A story can be locally coherent without having causality when a bunch of events together are about the same general activity but in the wrong order or not motivated by something that happened earlier in the story. 
%A story can have soft causality but not local coherence if a character is interleaving actions from different goals (picking up groceries while on the way to rob a bank).

%Commonsense reasoning is predicated on a shared set of beliefs about the dynamics of the world.
Commonsense knowledge is the set of commonly shared knowledge about how the world works. 
It enables us to form expectations about what will happen if we take certain courses of action and to infer things that likely happened in the past.
Commonsense reasoning is the application of commonsense knowledge to specific contexts.
Relevant to our work, commonsense reasoning might be applied to make inferences about what might have needed to have taken place for a character to arrive at a certain state---soft enablement---and what a reasonable next action would be based on what has happened so far---soft goal hierarchies.

Specifically for this paper, we use COMET~\cite{Bosselut2019COMETCT} to model an assumed reader's commonsense knowledge.
COMET is a transformer-based language model designed for commonsense inference and is trained on ATOMIC~\cite{sap2019atomic}.
ATOMIC is a dataset containing 877k instances of information relevant for everyday commonsense reasoning in the form of typed if-then relations with variables.
ATOMIC is organized into different relation types such as ``needs'', ``wants'', ``attributes'', and ``effects''. 
We specifically use the relations for ``wants'' and ''needs''.
An example of a cause using the {\em wants} relation is as follows, ``if X tried to get away, then X {\em wants} to be free.''
Likewise, an example of an effect using the {\em needs} relation is, ``if X scaled the wall, then X {\em needs} to know how to scale the wall.''

The key difference between hard and soft causality is the idea of expectations of causality via commonsense reasoning and can be illustrated using the relations seen here. 
A hard causal relation requires verification and satisfaction of propositions, as in the example given in the paper - John cannot shoot Xavier if John is not in possession of a gun or they are not co-located.
A soft causal relation here would be that the reader's belief that John dislikes Xavier and wants to fight him and thus as a result, he {\em wants} a weapon.
Guns are weapons and thus there is a probability that John {\em needs} a weapon to fight Xavier.

In the next section we detail how we use the theory of soft causal relations, and COMET commonsense inferences about needs and wants, to generate stories. 
In section 5, we present the results of a human participant study that uses an evaluation of several systems in two distinct genres to probe how soft causal relations affect participant perceptions of story quality and coherence.

\section{C2PO}
\label{sec:c2po}

This section presents the overall layout of \oursys{}.
\oursys{} works by first extracting a set of high level plot points from a given textual story plot $S$ and then generating a branching set of events that go between each high level plot point.
The final story is obtained by walking the overall plot graph generated by joining each generated sub-graph.

\subsection{Plot Extraction}
The overall plot extraction process is described in Figure~\ref{fig:extraction}.
In order to facilitate plot extraction, we propose a method that uses coreference resolution and information extraction to identify a set of plot points following a single character. % in a manner similar to the RAKE algorithm for automated keyword extraction~\cite{rake}
First, we extract all the coreference clusters using a pre-trained neural coreference resolution model \cite{Clark2016DeepRL}. 
There can be multiple such clusters, each of which contains all mentions in the story belonging to a single possible character.
We pick one of these clusters at random. 
Let $M=\{m_1,m_2,..., m_n\}$ denote this cluster.
%Each mention in the cluster often consists of a single word.
%Although one could potentially use the same coreference model to select an entire span of words as the context for a mention, we found that it would often lead to bloated outputs and unfinished sentences.
Simultaneously, we also extract a set $\mathcal{R}$ of $\langle subject,relation,object\rangle$ triples from the story text using OpenIE~\cite{Angeli2015}.%, this set is represented as $G=(V,R)$.
%To prevent this, we propose to instead use relational tuples obtained from OpenIE \cite{} over a mention's context. Not only does this yield more concise events, but an extracted tuple comes in three parts: subject, relation, and object phrases; information that we can use to further augment our method as well as help the next step in the process.

Once we have both of the set of mentions for a character and the triples for the story, we align them, attempting to find the subset of triples 
$\mathcal{P} \subset \mathcal{R}$ 
that are relevant for a single character on the basis of their character-level positions within the original story text.
Both the neural coreference model and OpenIE are information retrieval systems and so we can identify the character-level offset or position of the retrieved information in the original story text.
Let $pos(\cdot)$ be a function that can do this.
The set of plot points is $\mathcal{P}=\{\langle s,r,o\rangle:pos(m)=pos(s),\forall m\in M, \langle s,r,o \rangle \in G\}$.
The result is a sequence of relational tuples in which the character is the primary subject of the triple, ordered by when they first appeared in the original story text.
Joining each triple together yields a subject-relation-object phrase which we refer to as a \textit{plot point}.
%After extraction, we move on to alignment. For each mention in a candidate coreference cluster, we can match the reference with a corresponding relational tuple. This is done by comparing the character-level offsets of the coreference mention and the relation's subject and checking if there are any overlap within plot text which they are extracted from. The result is a sequence of relational tuples in which the character is the primary subject of the relation. Joining the each tuple triple together yields a subject-relation-object phrase which we refer to from here on as a \textit{plot point}.
\begin{figure*}[t]
    \centering
    \includegraphics[width=.865\linewidth]{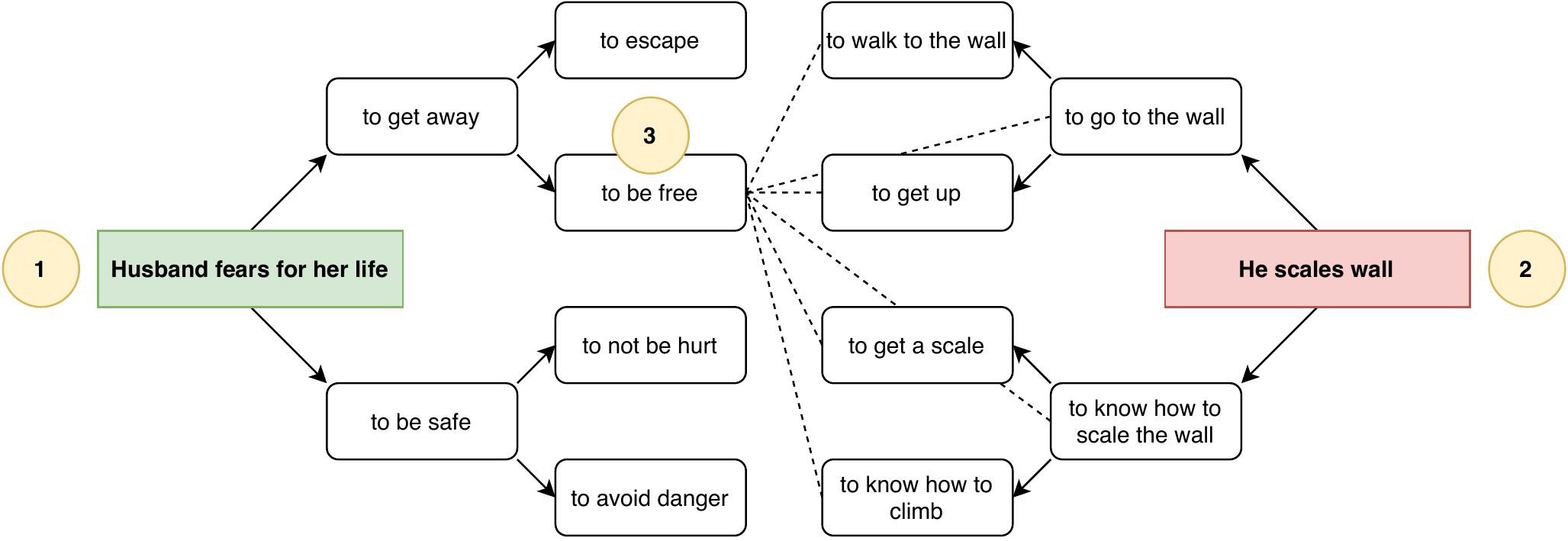}
    \caption{A demonstration of the plot graph generation process. 1 and 2 respectively indicate adjacent, extracted plot points. Dotted lines represent the process of finding the optimal link between the backward plot graph and node 3.}
    \label{fig:plotgraph}
\end{figure*}
\subsection{Plot Graph Generation}

Once we have established a series of plot points $\mathcal{P}=\{p_1,p_2,..., p_n\}$, we move on to plot graph generation as illustrated in Figure~\ref{fig:plotgraph}.
A plot graph is generated for each pair of adjacent plot points $(p_i, p_{i+1}),i\in\{1,..,n-1\}$ and then linked together in the order the plot points first appear in $P$ to form a plot graph for an entire story.

The process to generate a plot graph between adjacent plot points $p_1, p_2$ is as follows.
Starting from $p_1$, we use COMET~\cite{Bosselut2019COMETCT} to generate candidate next events in the story.
% COMET is a transformer-based language model designed for commonsense inference and is trained on ATOMIC~\cite{sap2019atomic}.
% ATOMIC is a dataset containing 877k instances of information relevant for everyday commonsense reasoning in the form of typed if-then relations with variables.
% ATOMIC is organized into different relation types such as \Mark{X, Y, Z, needs, wants}. 
% %
% We specifically use the relations for ``wants'' and ''needs''.
% An example of a cause using the {\em wants} relation is as follows, ``if X tried to get away, then X {\em wants} to be free.''
% Likewise, an example of an effect using the {\em needs} relation is, ``if X scaled the wall, then X {\em needs} to know how to scale the wall.''
%directly corresponding to forward causal establishment---a character has a want and therefore performs an action---and backward enablement---a character needed something to be true to do an action.
The {\em wants} relation indicates a direct forward cause---a character has a want and therefore performs an action.
%\footnote{By focusing on character-centric stories, we do not model causal effects by non-intention-possessing entities such as physics and nature (e.g., gravity causes a ball to roll). For an analysis of the relationship between causality and character goals see~\citet{graesser-trabasso}.}
We recursively query COMET to generate $k$ event candidates $n$ times going forward starting with $p_1$; let this be $\mathcal{G}^f$.
The {\em needs} relation indicates backward enablement---a character needed something to be true to do an action.
We recursively query COMET to generate $k$ event candidate $n$ times going backward from $p_2$; let this be $\mathcal{G}^b$.
This gives us two directed acyclic graphs as seen in Figure~\ref{fig:plotgraph}.
The relations in $\mathcal{G}^f$ and $\mathcal{G}^b$ are weighted proportional to the likelihood score produced by COMET for each inference.

The next step is to look for the optimal way to link $\mathcal{G}^f$ and $\mathcal{G}^b$ and computing the probability of reaching a node $u\in \mathcal{G}^f$ looking at all nodes $\forall v \in \mathcal{G}^b$.
Let
%$v_f\in P_f$ and $v_b\in P_b$,
$Pr^{needs}(u|v)$
be the probability of generating event $e_2$ as determined by COMET under the $needs$ relation, conditioned on $e_1$, 
and $Pr^{wants}(v|u)$ be the same but under the $wants$ relation. 
We define this link's weight as:
\begin{equation}
w(u,v) = \frac{Pr^{wants}(u|v)}{\alpha^{wants}_u} + \frac{Pr^{needs}(v|u)}{\alpha^{needs}_v} 
\end{equation}
were $\alpha^{wants}_u$ and  $\alpha^{needs}_v$ are normalization constants. 
Here we set them equal to the probability of generating the word ``to'', a word in ATOMIC common to both relation types.
This process is repeated for all nodes until we have found a set of optimal links.~\footnote{COMET and ATOMIC can be replaced by any model designed for automated knowledge base completion and corresponding commonsense reasoning knowledge base by selecting the appropriate relations in the replacements.} 

Finally, we link together the plot graphs for the entire sequence of plot points: $\mathcal{G}=\bigcup_{p_1,p_2}(\mathcal{G}^f_{p_1}\cup\mathcal{G}^b_{p_2}), \forall p_1,p_2\in \mathcal{P}$ where $p_1,p_2$ are adjacent in $\mathcal{P}$. 
%for all pairs of plot points $p_1, p_2\in\mathcal{P}$ 
A story can be generated via a random walk of the graph from the first plot point $p_1$ to the last $p_n$.
All random walks are guaranteed to terminate in $p_n$ because $\mathcal{G}^b_{p_n}$ is constructed by branching backward from $p_n$.
Likewise, each intermediate plot point $p_2...p_{n-1}$ is a node in $\mathcal{G}$ that all walks must pass through.

%we integrate COMET \cite{}, a transformer-based language model used for common-sense inference. Specifically, we leverage the models which model the wants and needs relations, denoted by $xWant$ and $xNeed$.

%For each pair of adjacent plot points, denote the starting point be $S_0$ and ending point be $E_0$. To get candidates for our next We start by generating using the $xWant$
%\Wes{Forward and backward gen, xNeeds relation meets xWant}

%\TODO{plot graph reintro} 
%Nodes on the graph that occur earlier when grouped by levels in the topological ordering are considered dependencies required for later events to occur. 
%\TODO{math}
\section{Experiments}
%2 Genres - mystery and Fairy
%dataset from bringing stories alive paper

\begin{table}[]
\centering
\footnotesize
\begin{tabular}{l|l|l}
                    & \textbf{Mystery} & \textbf{Fairy Tale} \\ \hline
No. Stories         & 569                   & 695                    \\
Sentences per story & 23.36                   & 24.80                    \\
Vocabulary size     & 21,238                   & 16,452                    \\
%Unique words        &                    &                    
\end{tabular}
\caption{Dataset statistics.}
\label{tab:data}
\end{table}

\begin{table}
\centering
\footnotesize
\begin{tabular}{l|c|c}
                      & \textbf{Commonsense} & \textbf{Storytelling} \\ \hline
\textbf{C2PO}         &             \checkmark                  &          \checkmark                     \\
\textbf{BERT+infill}    &             \checkmark                  &                               \\
\textbf{Hier. Fusion} &                               &                     \checkmark         
\end{tabular}
\caption{Inductive biases of each system.}
\label{tab:modelcriteria}
\end{table}

\begin{table*}
\scriptsize
\centering
\begin{tabular}{p{1cm}|p{4.1cm}|p{4.1cm}|p{5.1cm}}
                    & \multicolumn{1}{c|}{\textbf{C2PO}} & \multicolumn{1}{c|}{\textbf{BERT+infill}} & \multicolumn{1}{c}{\textbf{Hierarchical Fusion}} \\ \hline
\textbf{Mystery}    
&       \textbf{Holmes decides go.} Holmes wants to go. Holmes begins to see something. Holmes begins to look around. \textbf{Holmes notices a pair of trouser knees.} Holmes wants to clean up. Holmes begins take a shower. Holmes wants to get ready. Holmes wants to walk to the store. \textbf{Holmes taps in front of Wilson's shop.} Holmes tries say hello. Holmes wants start the car. Holmes tries to drive to the scene. \textbf{He calls Police Inspector Jones.}        
&        \textbf{Holmes decides go.} Holmes new friend initially stays. Holmes new son accepts goes. Holmes mother also stays.  \textbf{Holmes notices a pair of trouser knees.} Holmes himself still watches. Holmes again is house ghost watches. Holmes insists he took watch. \textbf{Holmes taps in front of Wilson's shop.} Holmes smiles and eventually leaves. Holmes red cap now appears. Holmes silhouette finally stands. \textbf{He calls Police Inspector Jones.}           
&         \textbf{Holmes decides go.} The room was silent. The room was silent. The air was heavy , and the room was quiet.  \textbf{Holmes notices a pair of trouser knees.} The young man wasn't going to be a father. His parents weren't supposed to be a father. They had the best kids in the entire world. \textbf{Holmes taps in front of Wilson's shop.} I'm not sure what's happening to me , but I'm not sure. What? You've been in a heel for a few years, and you've been in a heel for nearly a month. \textbf{He calls Police Inspector Jones.}                      \\ \hline
\textbf{Fairy Tale} 
&      \textbf{Queen asks her mirror.} Queen wants to look better. Queen wants to try on clothes. Queen starts to be mad. \textbf{Queen is furious.} Queen tries to relax. Queen wants to take a nap. Queen starts to get up. Queen begins to approach someone. \textbf{She appears at a dwarfs'.} Queen starts to surprise everyone. Queen starts to have a party. queen wants to have money. Queen tries to buy poison comb. \textbf{She brushes with poisoned comb.} Queen tries to wash her hair. Queen starts dry it. Queen wants to be hungry. Queen wants to get the knife. \textbf{Queen cuts the apple in half.}        
&     \textbf{Queen asks her mirror.} Queen is still half smiles. Queen who had had frowns. Queen has always asked. \textbf{Queen is furious.} Queen wife of mary then flees. Queen wife husband anna maria refuses. Queen mistress queen mistress wives demands. \textbf{She appears at a dwarfs'.} Queen queen rose meets princess. Queen sees fairies she crowns fairies. Queen rises with beauty. \textbf{She brushes with poisoned comb.} Queen was now also finally returns. Queen then had only disappears. Queen thought she vanished. \textbf{Queen cuts the apple in half.}
&     \textbf{Queen asks her mirror.} ``What the ...'' ``You know I have no idea how I got here. You know I can't do anything about it.'' ``I know I can't do anything about it.'' \textbf{Queen is furious.} A large, creature sits in the middle of a room with an odd looking cat on it. The creature is a strange looking cat, though it looks like the same cat is in its own room. Its fur is like a large, white slept. \textbf{She appears at a dwarfs'.} ``So, you 're here to kill me,'' asked the man in the suit, with a slight hint of worry in his due. ``Yes,'' replied the man in the suit. \textbf{She brushes with poisoned comb.} We hadn't met in a long time. We weren't supposed to be alone , and the rest of our group was just a group of people. \textbf{Queen cuts the apple in half.}           
\end{tabular}
\caption{Examples of a story generated by each model in both genres given the same initial set of {\em bolded} high level plot points. Further randomly selected examples can be found in Appendix~\ref{sec:supplemental}}
\label{tab:examples}
\end{table*}
%The overall setup of the experiments is as follows.
We evaluate on a story dataset with two genres---mystery stories and fairy tales---first introduced by~\citet{ammanabrolu20world}\footnote{\url{https://github.com/rajammanabrolu/WorldGeneration}}, statistics for the dataset can be found in Table~\ref{tab:data}.
The data is partitioned into train and test splits in a 8:2 ratio, and the train split used to train \oursys{} and two baseline models (described below).
%The following procedure is repeated for each of these genres.
A random set of 10 stories is chosen from each genre in the test set and high level plot points are extracted as described in Section~\ref{sec:c2po}.
For each model and for each set of high-level plot points and for each genre we generate three distinct stories for a total of ($3\times 10 \times 2 \times 3 = 180$ stories.
% For each set of high level plot points and across the models we are comparing, we generate three stories.
We generate three stories for each combination of model, plot point set, and genre
to account for variance in stories that can be produced by the same high level plot due to the branching nature of \oursys{} as well as variance in the baselines' outputs.
Standard automated language generation metrics such as perplexity and BLEU~\citep{papineni-etal-2002-bleu} are known to be unreliable for creative generation tasks~\cite{ammanabrolu2019story}.
The stories are thus evaluated using a human participant study, described below.
%Details regarding the baselines and human participant study are given below.

\subsection{Baselines}
We choose two baselines on the basis of the comparisons they afford (summarized  in Table~\ref{tab:modelcriteria}).
Both are designed to perform text infilling tasks but differ based in their inductive biases.
%We choose models based on their inductive biases.
``Inductive biases'' here specifically refer to a system's ability to model commonsense knowledge and if they were originally designed for storytelling or not.
% The first is a BERT~\cite{devlin2018bert} based model that has not strictly been designed for storytelling but adapted to text infilling.
% %; BERT is known to contain implicit factual commonsense knowledge~\cite{petroni2019language}.
% The second is the Hierarchical Fusion model~\cite{Fan2018} that is designed for storytelling but does not explicitly model causality nor commonsense reasoning.
% \oursys{} is designed for storytelling uses commonsense reasoning to infer soft causal relations (Section~\ref{sec:causal}) via COMET's commonsense inference abilities.
% %\oursys{} explicitly utilizes commonsense knowledge via COMET and contains an inductive bias for story generation in the form of plot graphs.

\subsubsection{BERT+infill} 

The first baseline is a BERT~\cite{devlin2018bert} based model that has not strictly been designed for storytelling (though BERT is trained on a corpus that includes story texts) and then adapted to perform text infilling.
%BERT has been shown to contain implicit factual commonsense knowledge~\cite{petroni2019language}.
Although large-scale pre-trained language models are known not to be great storytellers, mostly due to them being unable to stay on track for any extended period of time~\cite{see-etal-2019-massively}, they have demonstrated knowledge of factual commonsense information by virtue of the amount of data they have been trained on~\cite{petroni2019language}.
Our problem setting requires us to generate a section of text between two consecutive high level plot points at a time, reminiscent of approaches taken by \citet{ippolito-etal-2019-unsupervised} and \citet{donahue2020enabling} that condition a language model on left and right contexts to fill in blanks in a story.
We follow a similar setup for this baseline, using BERT~\cite{devlin2018bert} conditioned to attend to both previous tokens---the preceding plot point---and future tokens---the following plot point---to generate sequences~\cite{lawrence-etal-2019-attending}.
BERT+infill is fine-tuned using this methodology on the high-level plot points extracted from our training data.
Despite being similar to these prior methods, we note that BERT+infill utilizes no storytelling domain knowledge in its architecture and boils down to simple masked language modeling with multiple mask tokens.
%\Wes{add any extra details}

\subsubsection{Hierarchical Fusion}

\citet{Fan2018} train their system---consisting of a convolutional sequence-to-sequence network with self-attention~\cite{ott2019fairseq}---on the Reddit Writing Prompt corpus, where human-contributed prompts are paired with human-contributed stories.
The system learns to first generate a prompt and then transform it into a story.
This model's architecture is explicitly designed to tell stories and is suited for a type of storytelling wherein a prompt for a story is generated into a passage
%---thus incorporating a form of storytelling inductive bias.
This type of training is particularly well suited to our setup of generating a story piece-by-piece using extracted high level plot points.
We train the model from our training set using high level plots extracted from the stories as described in Section~\ref{sec:c2po} as the prompts and sections in between each of these extracted plot points as the story.
%\Wes{add any extra details}

\begin{table}
\footnotesize
\begin{tabular}{l|ll|l}
                    & \textbf{C2PO vs. BERT+} & \textbf{C2PO vs. Hier.} & \textbf{Tot.} \\ \hline
\textbf{Mystery}    & 82                          & 89                             & 171            \\
\textbf{Fairy Tale} & 90                          & 90                             & 180            \\ \hline
\textbf{Total}      & 172                         & 179                            & 351           
\end{tabular}
\caption{Participant count statistics.}
\label{tab:participants}
\end{table}

\subsection{Human Evaluation Setup} %cite purdy

We have 10 sets of high level plots per genre and three generated stories per each plot for each of the models.
We recruited $351$ human participants via Mechanical Turk.
Criteria for enrollment included: (a)~fluency in English, and (b)~demonstrating an understanding of commonsense based causality in stories.
To screen participants for the latter we asked them to predict potential next events that could reasonably occur given a story scenario.
An example of such a question asked can be found in Appendix~\ref{sec:appendix}.

Human participants are given one story generated by \oursys{} and another evenly randomly picked from those generated by either BERT+infill or Hierarchical Fusion for the same plot.
The order that these stories are presented in is randomized to account for bias induced due to the ordering effect~\cite{waysofknowing}.
Each story pairing is seen by at least three participants.
Participant count statistics are given in Table~\ref{tab:participants}.

%There were 171 total participants for the mystery genre, with 82 in the \oursys{} vs. BERT+infill and 89 in \oursys{} vs. Hierarchical Fusion categories.
%Similarly, 
\begin{figure*}[t]
    \centering
    \begin{subfigure}{0.49\textwidth}
    \centering
    \includegraphics[width=\linewidth]{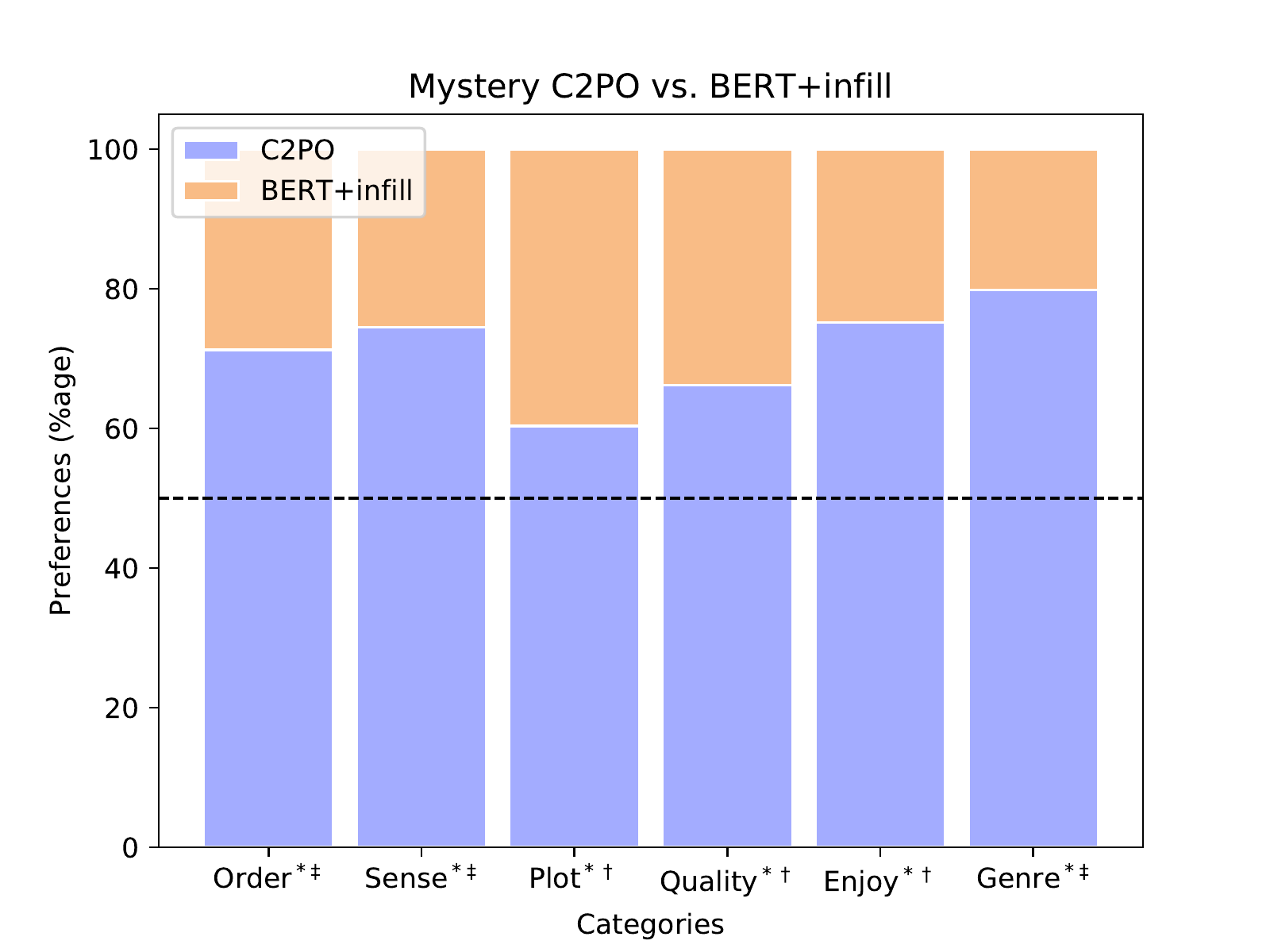}
    \caption{\oursys{} vs BERT+infill in the mystery genre.}
    \label{fig:resmystbert}
    \end{subfigure}
    \begin{subfigure}{0.49\textwidth}
    \centering
    \includegraphics[width=\linewidth]{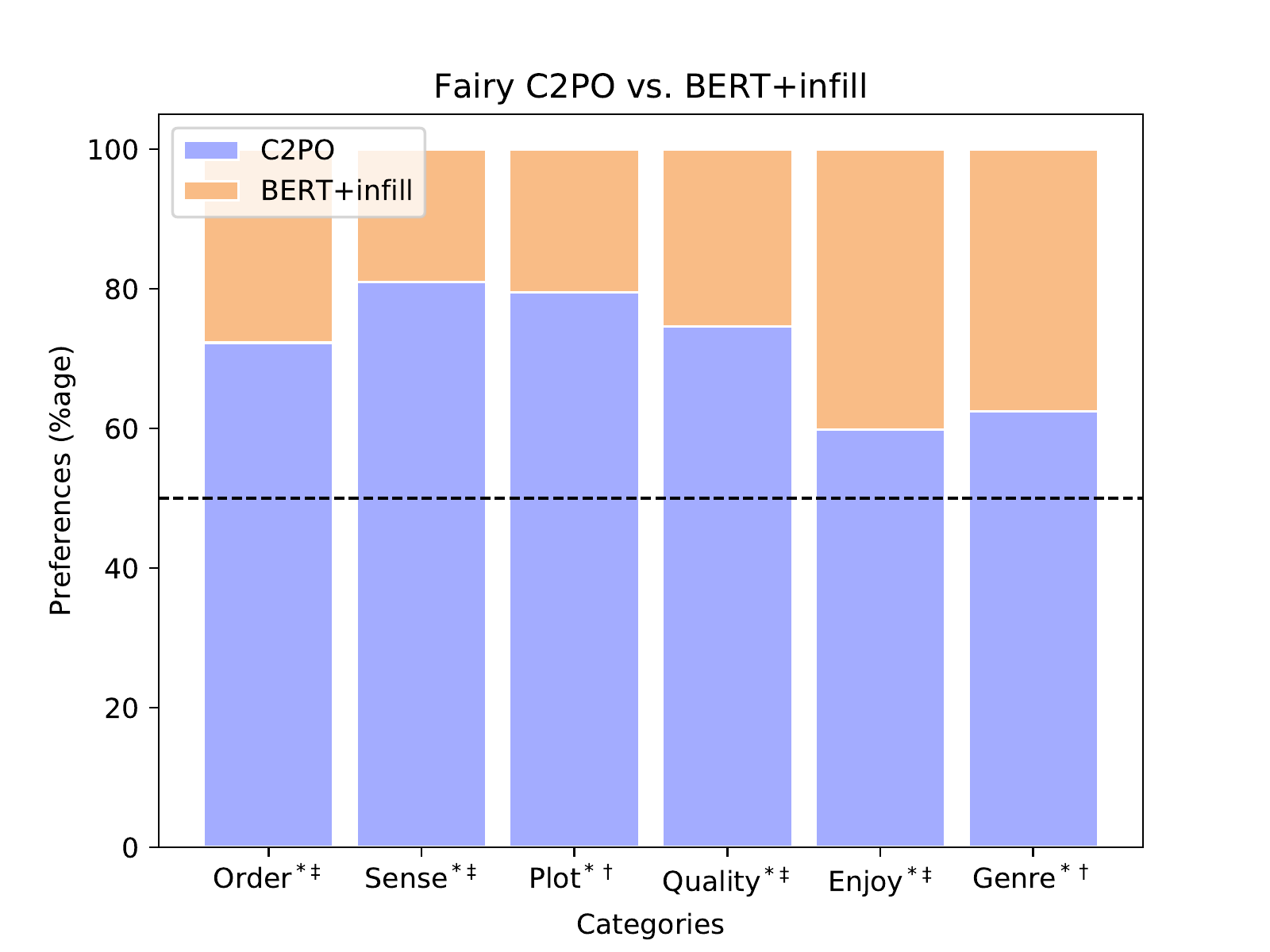}
    \caption{\oursys{} vs BERT+infill in the fairy genre.}
    \label{fig:resfairbert}
    \end{subfigure}
    \caption{Human evaluation results comparing \oursys{} vs BERT+infill. $^*$ indicates $p<0.05$, $^{\ddagger}$ indicates $\kappa>0.4$ or moderate agreement, $^\dagger$ indicates $\kappa>0.2$ or fair agreement}
    \label{fig:resultbert}
\end{figure*}

Participants are then asked a series of questions, each measuring a particular aspect of perceived story quality, comparing the \oursys{} generated model to one of the baselines.
For each question they are asked to note down which story they preferred.
The questions we use are adapted from \citet{purdy2018predicting} and have been used in multiple storytelling works as an indication of story quality~\cite{Tambwekar2019,ammanabrolu2019story}.
Specifically, we ask:
\begin{itemize}[noitemsep]
    \item Which story’s events occur in a more PLAUSIBLE ORDER?: as a proxy to indicate perceptions of overall causality within the story.
    \item Which story’s sentences MAKE MORE SENSE given sentences before and after them?: to examine perceptions of local causality and commonsense reasoning in the story.
    \item Which story better follows a SINGLE PLOT?: for insight into perceptions of global coherence for the entire story.
    \item Which story is of HIGHER QUALITY?: as a measure of overall perceived story quality.
    \item Which story is more ENJOYABLE?: indicates story value.
    \item Which story better FITS A GENRE?: as a measure of how well the story matches commonsense knowledge specific to a genre, capturing the differences between the two genres.
\end{itemize}

\noindent
For each of these questions, within a pairwise comparison, we perform a paired Mann-Whitney U test to assess statistical significance and additionally calculate Fleiss' $\kappa$ (Kappa) value to measure inter-rater reliability.

\section{Results and Analysis}
\begin{table}[]
\scriptsize
\begin{tabular}{l|ll|ll|ll}
                & \multicolumn{2}{l|}{\textbf{C2PO}} & \multicolumn{2}{l|}{\textbf{BERT+infill}} & \multicolumn{2}{l}{\textbf{Hierarchical}} \\ \hline
                & Myst,          & Fairy          & Myst,             & Fairy            & Myst,           & Fairy          \\ \hline
Avg. Sent/Story & 29.23                 & 30.2               & 25.4                    & 26.0                 & 31.3                  & 41.0               \\
Avg. Words/Sent & 4.94                 & 5.04               & 4.62                    & 4.79                 & 7.21                  & 5.75               \\
Unique Bigrams  & 312                 & 317                & 356                    & 357                 & 380                   & 402               \\
Unique Trigrams & 1245                 & 1353               & 1856                    & 1870                 & 2187                  & 2190              
\end{tabular}
\caption{Statistics for generated stories. Unique n-grams are measured with respect to those found in the test set of the initial story data.}
\label{tab:generatedstats}
\end{table}

%\TODO{describe results and stat tests done on them and stats of actual generated stories}

There are a few dimensions along which we will attempt to analyze these results: (1)~the inherent inductive biases of each model as seen in Table~\ref{tab:modelcriteria}, (2)~the two genres, and (3)~the questions asked of the participants.
The analysis will be performed hierarchically in the order just presented.
% Figure~\ref{fig:resultbert} summarizes results comparing \oursys{} to BERT+infill for both genres, across all questions.
% Similarly, Figure~\ref{fig:resultshier} compares \oursys{} and the Hierarchical Fusion model.
Table~\ref{tab:generatedstats} provides statistics on generated stories and Table~\ref{tab:examples} displays select examples of generated stories for each of the models in both genres.
% Examples of stories and statistics can be found in Appendix~\ref{sec:samplestories}\TODO{}.

\begin{figure*}[t]
    \centering
    \begin{subfigure}{0.49\textwidth}
    \centering
    \includegraphics[width=\linewidth]{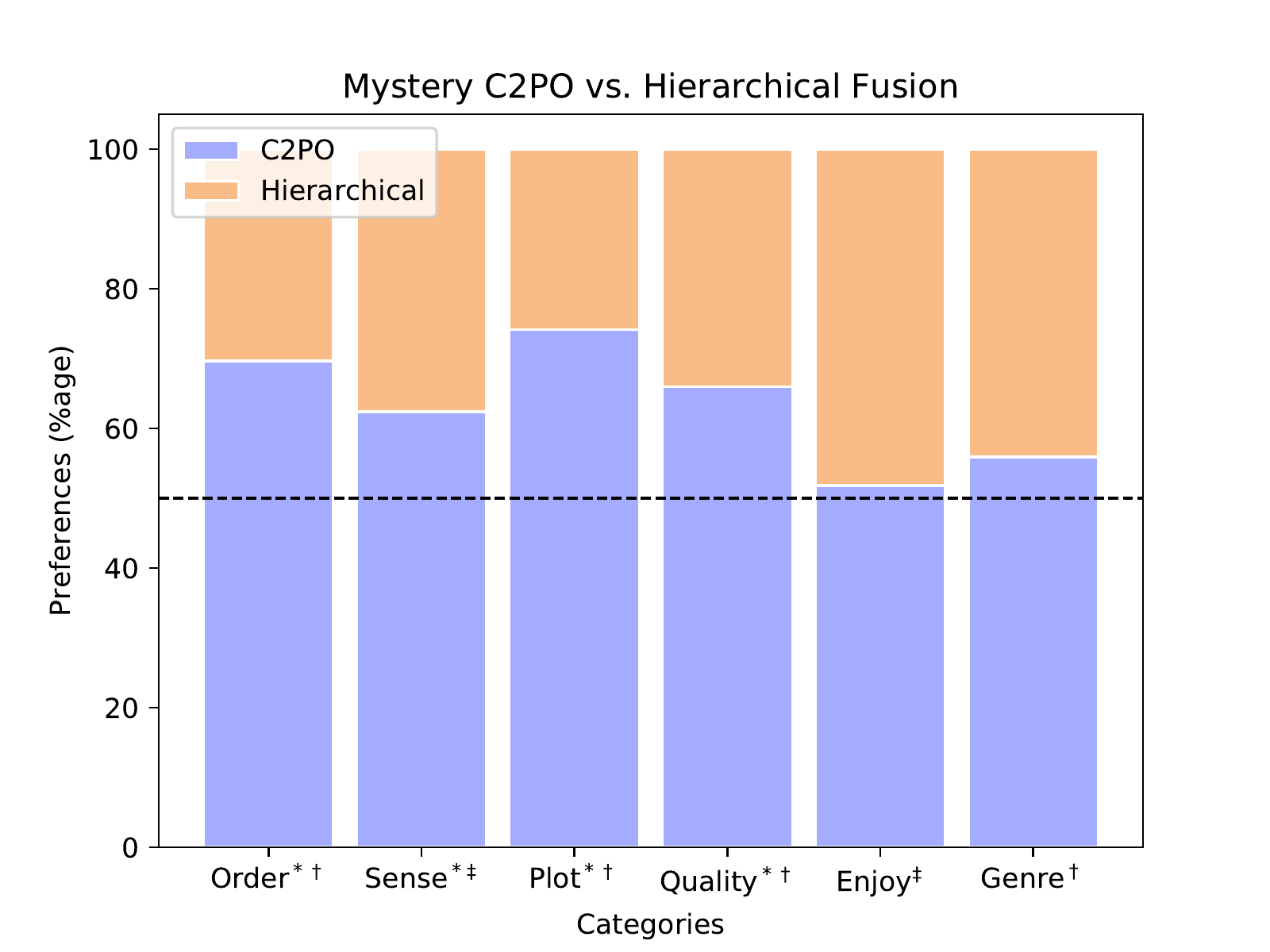}
    \caption{\oursys{} vs. Hierarchical Fusion in the mystery genre.}
    \label{fig:resmysthier}
    \end{subfigure}
    \begin{subfigure}{0.49\textwidth}
    \centering
    \includegraphics[width=\linewidth]{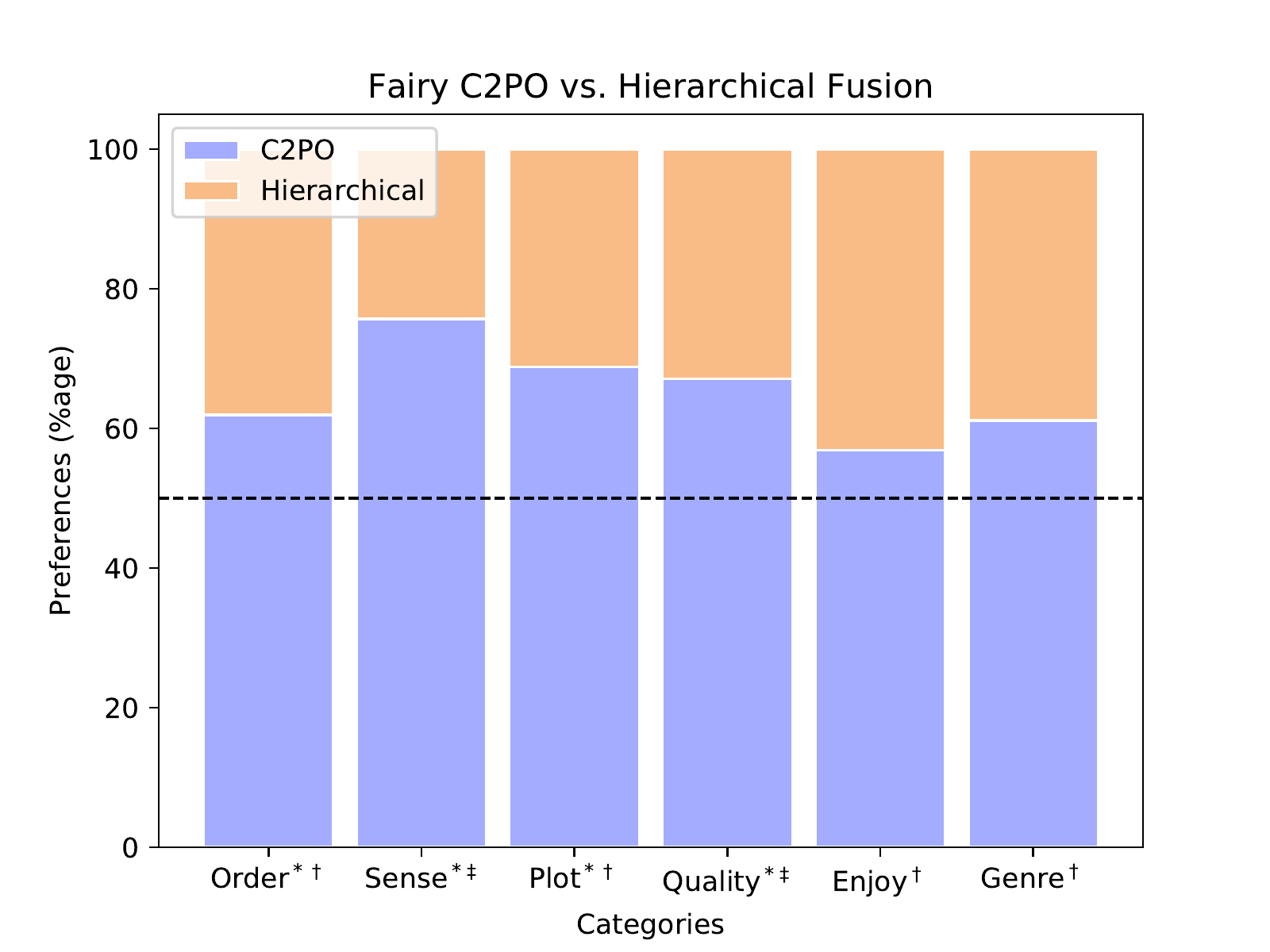}
    \caption{\oursys{} vs. Hierarchical Fusion in the fairy genre.}
    \label{fig:resfairhier}
    \end{subfigure}
    \caption{Human evaluation results comparing \oursys{} vs, Hierarchical Fusion. $^*$ indicates $p<0.05$, $^{\ddagger}$ indicates $\kappa>0.4$ or moderate agreement, $^\dagger$ indicates $\kappa>0.2$ or fair agreement}
    \label{fig:resultshier}
\end{figure*}

\subsection{\oursys{} vs BERT+infill}

Figures~\ref{fig:resmystbert} and  \ref{fig:resfairbert} show the percentages that participants preferred \oursys{} versus the BERT+infill system for each dimension and for each story genre.
\oursys{} is preferred over BERT+infill in both genres and in all dimensions. 
All of these results are statistically significant ($p < 0.05$) with fair-to-moderate inter-rater reliabilities.

For the mystery genre the greatest differences in preferences are observed with respect to enjoyability and genre resemblance.
The systems were most similar with regard to their ability to maintain a single plot.
%---likely due to having a shared set of plot points.
For the fairy tale genre
the greatest differences are seen in terms of the story events' plausible ordering, making sense causally, and the ability to maintain a single plot.
The models were most similar with regard to their genre resemblance and enjoyability.

The questions that \oursys{} does particularly well on compared to BERT+infill are complementary across the genres.
Enjoyability and genre resemblance are rated higher for \oursys{} in the mystery genre as opposed to fairy tales.
We additionally observe that these two factors are highly, positively correlated using Spearman's Rank Order Correlation ($r_s=0.56,p<0.01$).
%BERT+infill, simply finetuned on the data and 
Similarly, \oursys{} performed comparatively better in terms of plausible ordering, making sense causally, and the ability to maintain a single plot for fairy tales than for mysteries.
These three factors are also highly, positively correlated with each other and in terms of overall perceived story quality ($0.6>r_s>0.55$, $p<0.01$ for all pairwise comparisons).

This provides evidence that the brand of commonsense reasoning-based causality brought to bear by \oursys{}---needs and wants---works well in the mystery genre.
The mystery genre follows everyday commonsense norms whereas the fairy tale genre is more likely to stray from commonsense norms.
It can thus be inferred that genre-specific or thematic commonsense knowledge is required to improve perceptions of genre resemblance and enjoyability but does little in terms of metrics assessing local and global coherence in terms of causality.

\subsection{\oursys{} vs Hierarchical Fusion}

Figures~\ref{fig:resmysthier} and \ref{fig:resfairhier} show the percentages of participants that preferred \oursys{} to Hierarchical Fusion.
For the mystery genre, \oursys{} was preferred for the dimensions of plausible ordering, making causal sense, maintaining a single plot, and overall story quality.
These dimensions were significantly different ($p < 0.05$).
The dimensions of enjoyment and genre resemblance were not significantly different, meaning no system did better than the other.

We see a similar pattern for fairy tale stories: \oursys{} is preferred to hierarchical fusion for the same dimensions as the mystery genre and are not significantly different for enjoyment and genre resemblance.

Across genres, 
there is a positive correlation between metrics relating to coherence and overall perceived story quality ($0.6>r_s>0.5$, $p<0.05$ for each pairwise comparison using Spearman's Rank Order Correlation).
Also recall that the Hierarchical Fusion model contains an inductive bias for storytelling but does not model commonsense reasoning.
This appears to indicate that genre resemblance and enjoyability are not dependant on causal, commonsense reasoning but rather on the how much the generated text ``sounds like a story'' but story quality still depends on overall coherence.

\subsection{Broader Trends}
%positive correlation between making sense and ordering shows  first shows interdependency of causality and csense
%having csense modeling is good but works even better if you can capture thematically relevant info
%LMs by themselves have no penchant for storytelling
%having a storytelling inductive bias is good for improving genre resemblance and enjoyability but does little in terms of perceptions of local/global coherence in terms of causality and adherence to ccsense - which adversely affects overall perceived story quality
%whole is greater than the sum of the parts and we really need both a storytelling bias and csense reasoning for language generation systems to be able to tell stories - just off the shelf one or the other don't really work

%storytelling inductive bias basically has the thing sounding more like a story in terms of style as seen in table 5 but doesn't really do well in terms of the content
%only modeling commonsesne gives  you vice versa

There are two main trends that one can see across the models depending on their inductive biases (extent to which the models are trained for commonsense reasoning or storytelling).
%as seen in Table~\ref{tab:modelcriteria}.
We observe these trends on the basis of the analysis presented so far as well as the examples of output stories found in Table~\ref{tab:examples}.
(1)~Having commonsense reasoning abilities generally improves perceptions of local and global coherence in terms of causality with a caveat that what is perceived as commonsense can change across genres.
When genre or domain specific commonsense knowledge matches ``everyday'' commonsense, it makes for an automated storyteller that is significantly more causal in nature.
(2)~Just commonsense reasoning without any sort of storytelling inductive bias incorporated---such as with pre-trained and finetuned language models which themselves have no real penchant for storytelling---into a model's design doesn't help, however, in terms of enjoyability and genre resemblance.
The performance of Hierarchical Fusion in terms of enjoyability and genre resemblance---and the examples seen in Table~\ref{tab:examples}---appear to indicate that models designed for storytelling do a better job of maintaining the writing style of a story but struggle with causality.

%bunch of dimensions to compare across: what each model was designed for, the genres, the actual questions
%gonna compare at the top level in the order
%incorporate storytelling inductive bias?
%encourage future works to explicitly introduce inductive biases for both commonsense reasoning and for storytelling simultaneously

%what we've really found here is that commonsense and causality are correlated
%both of these are then also related to perceptions of quality
%keeping the extraction constant, we see less pronounced differences in results between mystery and fairy
%prob cause comet fails in the fairy domain
%what have we found is that actually thematic commonsense and causality are critical to perceptions of overall story quality in a particular genre
%commonsense shifts per genre and although there is some amount of commonsense reasoning that transfers and translated to enhanced perceptions of causality, in general you need to account for the fact that it is different
%but just commonsense by itself isn't exactly enough now is it, storytelling 

%now for mystery, c2po does better than baselines 
%we find that although commonsense doesnt exactly help it feel more like a fairy tale or increase enjoyability, it does strongly increase overall coherence - this may be in part due to the fact that the whole implicit commonsense thing for bert breaks down fully and loses coherence when confronted with a domain that breaks commonsense expectations; as far as fusion goes 

\section{Conclusions}
%\Mark{Probably needs to be shortened and rewritten.} check it out now
We intend for the findings of this work to be utilized by researchers studying automated storytelling, a standing AI grandchallenge requiring creative, long-form language generation.
%Most prior works in the area either incorporate inductive biases in the form of either commonsense reasoning abilities or storytelling domain knowledge and do not measure the impact of these biases across a wide set of human perceived metrics relating to story quality.
We explore the effects of {\em soft causal relations}---reasonable expectations by a reader regarding a story's progression---on human-based perceptions of overall story quality.
We introduce \oursys{} as a way to use {\em soft causal relations} via transformer-based models trained for commonsense inference in storytelling.
%which transformer-based models trained for commonsense inference~\cite{Bosselut2019COMETCT} with plot graphs---a structure that has been used extensively for storytelling systems in the past~\cite{Li}.

A key insight from a human participant study, measuring a wide set of human perceived metrics, shows that the sum of the parts is indeed greater than the whole.
Automated storytellers require both domain specific commonsense reasoning abilities as well as a storytelling inductive bias incorporated into the design of the system to perform well in terms of: local and global coherence on the basis of causality, enjoyability, genre resemblance, and overall story quality.
Further, perceptions of causal, commonsense conforming coherence are highly correlated with overall story quality.
We encourage authors of future work to build on these findings and more closely explore lines of research that use thematically relevant {\em soft causal relations} to improve automated storytellers. 

\section{Broader Impacts}
As an AI grandchallenge, automated storytelling consists of multiple sub-problems and thus has implications extending to the creation of learning agents that communicate effectively using natural language.
Our system faces the same potential pitfalls as other contemporary language learning systems. Namely, it is prone to echoing the biases present in the dataset~\citep{sheng-etal-2019-woman}.
Some of these biases are thematic and give the reader a sense of the genre.
{\em Prejudicial} biases are potentially more harmful.
Story generation can involve non-normative language use, describing situations that fictional characters engage in that would be considered inappropriate if enacted in the real world.
Crowdsourcing---in the case of ATOMIC~\citep{sap2019atomic} which COMET~\citep{Bosselut2019COMETCT} is trained on---and data curation---in the case of the story dataset used from~\citet{ammanabrolu20world}---can mitigate, but do not entirely eliminate these biases.
Story generation is broadly applicable (though not in its current state), from video game quests to conversational agents to book and movie generation.
As with any broad capability technology, it can be put to purposes that are benign, malicious, or negligent.
Fictional stories that are intentionally or unintentionally presented as true is a form of deception; we advise that future application developers using story generation technologies are very clear about the provenance of the story content delivered to an audience.

%\fontsize{9.0pt}{10.0pt}
%\selectfont
\bibliography{aaai21.bib}

\newpage

\section{Appendices}
\label{sec:appendix}
An example of a question used to screen participants on the basis of demonstrated everyday causal, commonsense reasoning:
\begin{enumerate}
    \item Select all of the following events that are likely to happen after the following story event: "The man ran into a solid brick wall with handholds."
    \begin{itemize}
        \item The man broke through the wall.
        \item The man climbed the wall.
        \item The man flew over the wall.
        \item The man made the wall disappear.
    \end{itemize}
\end{enumerate}
\subsection{Experiment parameter details}
\label{sec:hyperparams}
Hyperparameters for training and finetuning the BERT, COMET, and hierarchical models are consistent with their respective work's parameters, but we list them below where available. No additional hyperparameter tuning was conducted on any model.
\begin{center}
\begin{tabular}{ |c|c|c| } 
\hline
Parameter & BERT & COMET \\
\hline
embedding size & 768 & 768 \\ 
layers & 12 & 12 \\ 
batch size & 32 & 32 \\ 
\hline
learning rate & 0.00005 & 0.00025 \\ 
$\beta_1$ & 0.9 & 0.9 \\
$\beta_2$ & 0.999 & 0.999 \\
\hline
\end{tabular}
\end{center}
When generating branches for the plot graphs with \oursys{}, we generate 3 branches recursively for a length of 3 times.

\subsection{Example Generated Stories}
\label{sec:supplemental}
The next two tables provide stories by randomly selected plots from first the fairy tale, then the mystery genre.
%\begin{minipage}{\linewidth}
%\centering
%\scriptsize
%\begin{minipage}{\linewidth}

\begin{table*}
\centering
\scriptsize
\begin{tabular}{p{4.2cm}|p{4.1cm}|p{5.5cm}}
                    \multicolumn{1}{c|}{\textbf{C2PO}} & \multicolumn{1}{c|}{\textbf{BERT+infill}} & \multicolumn{1}{c}{\textbf{Hierarchical Fusion}} \\ \hline
\textbf{They live out at time seven years.} Bearskin village is just came. Bearskin and family wants country. Bearskin only takes river gives valley. Bearskin gave purse of gold. Bearskin brothers also agreed. Bearskin with senior chief heads agreed. Bearskin daughter among elder sisters agrees. Bearskin promised return in three years. Bearskin they had had promised. Bearskin daughter was sisters agreed. Bearskin daughter sister married my family. \textbf{Her sisters ridiculed her.} Bearskin also always once reappeared. Bearskin who always is later left. Bearskin also becomes a. \textbf{Bearskin found devil again At end of seven years.} Bearskin says and so says. Bearskin he can sing now sings. Bearskin it has told him cries. \textbf{He fulfill his promise.} Bearskin polish grease nail nails. Bearskin polish cut boot. Bearskin helps clean burn wood cuts. \textbf{Bearskin clip his nails.} Bearskin boots and mr. Bearskin boots nails and boot. Bearskin leather leather toe boot. \textbf{He is good.} Bearskin brother had still also stood. Bearskin looked and then said. Bearskin claimed it did. \textbf{Bearskin dropped his half of ring.} Bearskin shifted and he changed. Bearskin and slowly transformed. Bearskin so gently lifted and turned. \textbf{He was her bridegroom.}
& \textbf{They live out at time seven years.} Bearskin village is just came. Bearskin and family wants country. Bearskin only takes river gives valley. Bearskin gave purse of gold. Bearskin brothers also agreed. Bearskin with senior chief heads agreed. Bearskin daughter among elder sisters agrees. Bearskin promised return in three years. Bearskin they had had promised. Bearskin daughter was sisters agreed. Bearskin daughter sister married my family. \textbf{Her sisters ridiculed her.} Bearskin also always once reappeared. Bearskin who always is later left. Bearskin also becomes a. \textbf{Bearskin found devil again At end of seven years.} Bearskin says and so says. Bearskin he can sing now sings. Bearskin it has told him cries. \textbf{He fulfill his promise.} Bearskin polish grease nail nails. Bearskin polish cut boot. Bearskin helps clean burn wood cuts. \textbf{Bearskin clip his nails.} Bearskin boots and mr. Bearskin boots nails and boot. Bearskin leather leather toe boot. \textbf{He is good.} Bearskin brother had still also stood. Bearskin looked and then said. Bearskin claimed it did. \textbf{Bearskin dropped his half of ring.} Bearskin shifted and he changed. Bearskin and slowly transformed. Bearskin so gently lifted and turned. \textbf{He was her bridegroom.}

& \textbf{They live out at time seven years.} He was standing here in a large room. The door was wide open , and he looked down. He was n't sure what he was seeing , but he was n't sure what to think.Bearskin gave purse of gold. `` Hey , buddy , '' the man said , `` You 've got to be kidding me. '' The man laughed , `` Yeah , man. '' `` I 've been watching this guy for a long time , buddy.Bearskin promised return in three years. `` What do you mean I 'm here for you. '' `` Well , '' said the young boy , `` I 'll have you know this is n't a bad idea. '' `` But , '' said the young girl , `` I was n't in any trouble.\textbf{Her sisters ridiculed her.} `` What the hell. '' He was a man in a suit of black suit with a black tie. His face was red with sweat and a scar that was almost like lightning.\textbf{Bearskin found devil again At end of seven years.} `` Hey , man , what the Dom are you doing. '' I said as I looked around. `` I 'm not sure , man , '' he said with an expression of ease , `` you 're not really my man.\textbf{He fulfill his promise.} The man was n't exactly the most spinning man in the world. He was n't the most spinning man in the world. He was n't the most spinning man in the world , but he was n't the most famous man in the world.\textbf{Bearskin clip his nails.} I was n't the best on this wonders. I did n't need any. There is n't much time.\textbf{He is good.} `` So , what are you doing here. ''`` Oh , it 's been a while since you 've been in this business. ''`` I do n't really know much about this.\textbf{Bearskin dropped his half of ring.} `` Sir. This is the greatest news ever noble. '' He exclaimed as he glaring off to his office.\textbf{He was her bridegroom.} 
\\ \hline  
\textbf{Prince calls at time night.} Prince and tries again calls. Prince that still wakes. Prince then knocks and asleep. \textbf{He finds To his horror.} Prince robert francis charles george leaps. Prince charles henry louis rupert. Prince joseph john frederick maurice victor. \textbf{He leaps from tower.} Prince anthony or saint john leaves. Prince anthony nicholas edward lawrence. Prince edward nicholas james peter george. \textbf{He wanders For years.} Prince james edward thinks. Prince james john james. Prince alexander rupert james augustus george. \textbf{He hears again her voice.} Prince frederick leopold albert louis. Prince william william rupert. Prince augustus ernest william. \textbf{He leads their twins.}

& \textbf{Prince calls at time night.} Prince and tries again calls. Prince that still wakes. Prince then knocks and asleep. \textbf{He finds To his horror.} Prince robert francis charles george leaps. Prince charles henry louis rupert. Prince joseph john frederick maurice victor. \textbf{He leaps from tower.} Prince anthony or saint john leaves. Prince anthony nicholas edward lawrence. Prince edward nicholas james peter george. \textbf{He wanders For years.} Prince james edward thinks. Prince james john james. Prince alexander rupert james augustus george. \textbf{He hears again her voice.} Prince frederick leopold albert louis. Prince william william rupert. Prince augustus ernest william. \textbf{He leads their twins.}

& \textbf{Prince calls at time night.} He was sitting alone in the dark , surrounded by the many stars. He was alone , and this was it. He had been waiting for this moment , a long time ago.\textbf{He finds To his horror.} `` I 'm not really sure where I got that from , but I just want to know where I got it '' `` Oh , well it 's not a problem , you have to take it , just like I did '' `` But no it 's not , it 's not like you could just give it to me. I 'm not going to take it , but I do n't want to be like it. It 's not like you have any idea why '' `` You 're right , it 's not like you have the time.\textbf{He leaps from tower.} `` So ... '' `` What. '' `` I 'm not going to ask you any more questions about this. '' `` You 're going to be fine.\textbf{He wanders For years.} The sound of his phone being hung off the ground made him look up and see the phone. He had n't seen her since she was eight. She was the most beautiful thing he 'd ever seen.\textbf{He hears again her voice.} She 'd been sitting here for days. Her hair was long and dark , her hair a long and vice mess , a long , long scar across her forehead. Her hair was long , her hair long and vice.\textbf{He leads their twins.} 
\\ \hline  
\textbf{Girl has clothes.} Girl lives in dress. Girl know she can dress. Girl know still know she sees. \textbf{Her wear rags.} Girl not smile and talk. Girl never gets bad hair. Girl and most rarely smiles. \textbf{She do kinds of hard work.} Girl love to boy cry. Girl love girl get laugh. Girl do what will girls cry. \textbf{Girl go cry to God.}
& \textbf{Girl has clothes.} Girl lives in dress. Girl know she can dress. Girl know still know she sees. \textbf{Her wear rags.} Girl not smile and talk. Girl never gets bad hair. Girl and most rarely smiles. \textbf{She do kinds of hard work.} Girl love to boy cry. Girl love girl get laugh. Girl do what will girls cry. \textbf{Girl go cry to God.}
& \textbf{Girl has clothes.} The first few days were strike as hell. The first few days I woke up on a cold , damp night. My eyes were Small and my face looked as if I had been taller , bleeding , bleeding , I was just a little bit sprinting and confused by the pain.\textbf{Her wear rags.} The room was dark. Not a single light was on. All the lights in the room were on.\textbf{She do kinds of hard work.} The door opened , a dark figure in the dark. “ I am here to kill you , sir. I am here to kill you , sir.\textbf{Girl go cry to God.} 
\\ \hline  
\end{tabular}
\captionof{table}{Randomly selected examples of stories generated by the fairy models. Bolded sentences are the original extracted plot points.}
\label{tab:fairyexamples}
\end{table*}
%\end{minipage}

\begin{table*}
\scriptsize
\centering
\begin{tabular}{p{4.3cm}|p{4.1cm}|p{5.1cm}}
                    \multicolumn{1}{c|}{\textbf{C2PO}} & \multicolumn{1}{c|}{\textbf{BERT+infill}} & \multicolumn{1}{c}{\textbf{Hierarchical Fusion}} \\ \hline
\textbf{Holmes decides go.} Holmes wants to go. Holmes begins to see something. Holmes begins to look around. \textbf{Holmes notices has trouser knees.} Holmes wants to clean up. Holmes begins take a shower. Holmes wants to get ready. Holmes wants to walk to the store. \textbf{Holmes taps in front of Wilson 's shop.} Holmes tries to say hello. Holmes tries say hello. Holmes wants start the car. Holmes tries to drive to the scene. \textbf{He calls Police Inspector Jones.}

& \textbf{Holmes decides go.} Holmes new friend initially stays. Holmes new son accepts goes. Holmes mother also stays. \textbf{Holmes notices has trouser knees.} Holmes himself still watches. Holmes again is house ghost watches. Holmes insists he took watch. \textbf{Holmes taps in front of Wilson 's shop.} Holmes smiles and eventually leaves. Holmes red cap now appears. Holmes silhouette finally stands. \textbf{He calls Police Inspector Jones.}

& \textbf{Holmes decides go.} The room was silent. The room was silent. The air was heavy , and the room was quiet.\textbf{Holmes notices has trouser knees.} The young man was n't going to be a father. His parents were n't supposed to be a father. They had the best kids in the entire world.\textbf{Holmes taps in front of Wilson 's shop.} `` I 'm not sure what 's happening to me , but I 'm not sure. '' `` What. '' `` You 've been in a heel for a few years , and you 've been in a heel for nearly a month.\textbf{He calls Police Inspector Jones.} 
\\ \hline  
\textbf{Alec has room.} Alec tries to sleep. Alec wants take a shower. Alec begins to get dressed. Alec begins go to room. \textbf{Alec to room.} Alec tries to eat. Alec starts drink water. Alec wants to have money. Alec wants to have a car. \textbf{Alec throttle Holmes.} Alec starts to be successful. Alec tries to buy a car. Alec tries to go to the car. Alec wants to be in a car. \textbf{His father apparently twisting Holmes 's wrist.} Alec begins to hurt someone. Alec tries to do something bad. \textbf{Alec out of hand.}

& \textbf{Alec has room.} Alec back to says back says. Alec thinks back to goes. Alec well that sure did too. \textbf{Alec to room.} Alec sees and also sees baldwin. Alec himself sees baldwin waits. Alec herself eventually enters. \textbf{Alec throttle Holmes.} Alec fletcher holmes thomas john thomas. Alec watson james smith. Alec james stewart john hacking. \textbf{His father apparently twisting Holmes 's wrist.} Alec getting out suddenly went outside. Alec said i always hesitated. Alec really only half even laughed. \textbf{Alec out of hand.}

& \textbf{Alec has room.} I had been sitting in this room for a long time. I had never met a man before , but I had n't been there when I was in here. I was not sure why I was in here.\textbf{Alec to room.} `` What do you mean , it 's not real. ! '' `` Oh no. No.\textbf{Alec throttle Holmes.} This is not my first time writing. I 'm in a bit late for this so it 's not the first time I 've written anything but I 'm not going to start it. I hope I did n't mess up this.\textbf{His father apparently twisting Holmes 's wrist.} `` Hey you , '' said refuge. `` What. '' `` What 's this thing.\textbf{Alec out of hand.} 
\\ \hline  
\textbf{Brunton had deduced insofar ritual 's meaning.} Brunton tries to understand it. Brunton wants to learn. Brunton begins to look up. Brunton starts to look at trees. \textbf{He had determined elm tree 's height.} Brunton begins to take a break. Brunton starts to sleep. Brunton tries to get up. Brunton begins to approach someone. \textbf{He had had forced.} Brunton starts to get something. Brunton starts to buy it. Brunton starts to go to the store. Brunton starts to find out what they want. \textbf{He had chosen Rachel Howells.}
& \textbf{Brunton had deduced insofar ritual 's meaning.} Brunton then hadn fully understood. Brunton on first never thought. Brunton almost certainly knew. \textbf{He had determined elm tree 's height.} Brunton did then also stopped. Brunton saw d just immediately started. Brunton only still asked. \textbf{He had had forced.} Brunton two brothers too. Brunton wanted her twins. Brunton howell also wants twins. \textbf{He had chosen Rachel Howells.}

& \textbf{Brunton had deduced insofar ritual 's meaning.} A little early to this. I was on unseen and was n't quite sure about the first piece of writing. But I was n't really sure how to 15 to the story and I 'm not really sure how to improve .-- -- -He did n't seem like he should be.\textbf{He had determined elm tree 's height.} `` This is a joke , '' said the old man. `` It 's just a joke. '' `` I do n't believe it.\textbf{He had had forced.} `` This is the first time I 've had this problem. '' I said to myself as I stood up , my eyes wide and my head down with the realization of the situation I was in. I walked towards the door , my head down and the sound of a small wooden door being opened and a loud crack from the door echoed throughout the room.\textbf{He had chosen Rachel Howells.} 
\\ \hline  
\textbf{Wilder hired Hayes.} Wilder begins to give orders. Wilder wants to follow up. Wilder begins to hear news. \textbf{\textbf{Wilder heard news.}} Wilder tries to learn more. Wilder starts to do well. Wilder wants to work hard. He confessed all. Wilder begins to go home. Wilder begins to sleep. Wilder wants to get ready. Wilder begins to go to the restaurant. \textbf{He let his younger son stay at inn.} Wilder starts to go to bed. Wilder tries wake up. Wilder wants to work. Wilder begins to have money. \textbf{James Wilder seek his fortune.}

& \textbf{Wilder hired Hayes.} Wilder s brothers family initially agreed. Wilder s had resigned. Wilder sr announced d v. \textbf{\textbf{Wilder heard news.}} Wilder actually really cried. Wilder so alone has really wept. Wilder himself who only found sobs. He confessed all. Wilder and he refused. Wilder again is threatened. Wilder i again himself insisted. \textbf{He let his younger son stay at inn.} Wilder story was b. Wilder horror story by w. Wilder werewolf tale mr. \textbf{James Wilder seek his fortune.}

& \textbf{Wilder hired Hayes.} `` What is this. '' he asked , as he walked to the door. A door with a large metal door that was like an egg.\textbf{\textbf{Wilder heard news.}} The tree was still , the tree 's spirits was a tree 's tree. The tree was still , the tree , its tree and the tree were still. The tree was still , a tree , its tree and its tree and its tree.He confessed all. He walked into the bar and took a seat. He took a long , long drag of the cigarette. `` What have I done , '' he asked , `` You have to stop me , '' and he leaned forward to take another puff.\textbf{He let his younger son stay at inn.} The man looked at me and smiled. The man looked at me and said , `` You 're my only child , '' he said , `` I 'm sure your father is n't a bad man , '' he said. `` I do n't think I have the right to be like you , '' I said.\textbf{James Wilder seek his fortune.} 
\\ \hline  
\textbf{Colonel has behaviour.} Colonel begins to get better. Colonel wants to get up. Colonel starts to go to the door. Colonel wants to walk to the lock. \textbf{He would lock himself.} Colonel begins to get in the car. Colonel starts to drive. Colonel begins to drink. Colonel wants get drunk. \textbf{He shouting in drunken with pistol.} Colonel tries to sleep. Colonel begins to get up. Colonel wants to go outside. Colonel wants to go to garden. \textbf{He was found dead in garden pool.}

& \textbf{Colonel has behaviour.} Colonel was a must saw. Colonel is has did. Colonel not that was thought. \textbf{He would lock himself.} Colonel charles brown was. Colonel thomas and james a. Colonel thomas edward l. \textbf{He shouting in drunken with pistol.} Colonel general henry william miller killed. Colonel william andrew wilson acting. Colonel james edward richard stirling died. \textbf{He was found dead in garden pool.}

& \textbf{Colonel has behaviour.} `` You 're kidding me. '' I shouted. `` You 're joking about that.\textbf{He would lock himself.} `` I 'm sorry sir , but we do n't have the time. '' `` You did n't do this. '' `` We 're not here for that.\textbf{He shouting in drunken with pistol.} I 've been on this planet for three years. It 's been a few weeks , and it 's been quite some time since I 've been here. I 'm here.\textbf{He was found dead in garden pool.} 
\\ \hline  
\end{tabular}
\caption{Randomly selected examples of stories generated by the mystery models. Bolded sentences are the original extracted plot points.}
\label{tab:mysteryexamples}
\end{table*}
\end{document}